\documentclass[letterpaper, 10 pt, conference]{ieeeconf}  

\IEEEoverridecommandlockouts                              

\usepackage{caption}
\usepackage{comment}
\usepackage{lipsum}
\usepackage{color}
\usepackage{multirow}
\newcommand{\todo}[1]{{\color{red}\textbf{TODO} \textit{#1}}}
\newcommand{\etal}[1]{#1 et al.}
\usepackage{graphics}
\usepackage{graphicx}

\usepackage{amsmath}
\usepackage[ruled,vlined]{algorithm2e}
\usepackage{xcolor,colortbl}
\usepackage{nicefrac, xfrac}
\usepackage{amssymb}
\usepackage{subcaption}
\usepackage{float}
\usepackage{dblfloatfix}
\usepackage{cite}

\title{\LARGE \bf HD-OOD3D: Supervised and Unsupervised \\ Out-of-Distribution object detection in LiDAR data
}
\author{Louis Soum-Fontez$^{1}$ and Jean-Emmanuel Deschaud$^{1}$ and François Goulette$^{1,2}$
\thanks{$^{1}$Mines Paris, PSL University, Centre for robotics (CAOR), 75006 Paris, France; firstname.surname@minesparis.psl.eu}%
\thanks{$^{2}$U2IS, ENSTA, Institut Polytechnique de Paris, 91120 Palaiseau, France;
firstname.surname@ensta.fr%
}}

\begin{document}

\maketitle
\thispagestyle{empty}
\pagestyle{empty}


\begin{abstract}

Autonomous systems rely on accurate 3D object detection from LiDAR data, yet most detectors are limited to a predefined set of known classes, making them vulnerable to unexpected out-of-distribution (OOD) objects. In this work, we present HD-OOD3D, a novel two-stage method for detecting unknown objects. We demonstrate the superiority of two-stage approaches over single-stage methods, achieving more robust detection of unknown objects. Furthermore, we conduct an in-depth analysis of the standard evaluation protocol for OOD detection, revealing the critical impact of hyperparameter choices. To address the challenge of scaling the learning of unknown objects, we explore unsupervised training strategies 
to generate pseudo-labels for unknowns. Among the different approaches evaluated, our experiments show that top-K auto-labelling offers more promising performance compared to simple resizing techniques.
\end{abstract}


\section{INTRODUCTION}


LiDAR sensors, offering superior depth estimation and high-resolution 3D point cloud data, have become essential for autonomous vehicles, environmental mapping, and infrastructure monitoring, driving advances in 3D object detection as their availability and affordability increase.
Despite significant advancements, current neural network-based 3D object detection models still encounter substantial challenges when faced with "unknown" objects that were not part of the training dataset~\cite{openset3Dobj}. 

In real-world scenarios, encountering objects that lie outside the training distribution, referred to as out-of-distribution (OOD) objects, is inevitable and can pose significant safety risks. 
This issue is further exacerbated by the fact that the annotation of diverse object classes is both labor-intensive and costly, often leading to datasets that are not fully representative of real-world complexity. 

Specifically, traditional 3D object detection aims at outputting, for a given 3D LiDAR scan, $N$ bounding boxes represented as follows:
\begin{equation}
\{b_i = (cx_i, cy_i, cz_i, l_i, w_i, h_i, \theta_i, \hat{y})\}_{i \in [1, N] }
\end{equation}
where $cx_i, cy_i$, and $cz_i$ represent the center of the bounding box along the $x, y$ and $z$ coordinates,  $l_i, w_i$ and $h_i$ represent the length, width and height, and $\theta_i$ represents the heading of the bounding box, and $\hat{y}$ represents the predicted class amongst the $C$ annotated classes in the dataset. These bounding boxes should overlap with the ground truth bounding boxes.
In Open-Set 3D Object Detection, the output is similar, but now $\hat{y}$ represents the predicted class amongst the $C$ annotated classes in the dataset and an additional "unknown" class, representing all other relevant objects in the scenes. Another similar and relevant problem is Out-Of-Distribution (OOD) Detection, where from a set of predictions, an out-of-distribution measure is added:
\begin{equation}
\{b_i = (cx_i, cy_i, cz_i, l_i, w_i, h_i, \theta_i, \hat{y}, p_\text{OOD})\}_{i \in [1, N] }
\end{equation}

Works focusing on Open-Set 3D Detection suffer from a model performance degradation on known classes~\cite{openset3Dobj,Liang2023UnknownSF,towardsopen,mc_dropout}, which is why we frame the problem as an out-of-distribution task. The latter also preserves the original semantic information of the predicted object which is necessary for downstream tasks. 

A central component of our work is the development of a rigorous evaluation protocol. We show that key hyperparameters, such as distance thresholds and detection score cut-offs, have a substantial impact on the performance of OOD detection. By systematically analyzing these factors, we provide a more comprehensive and reproducible framework for evaluating OOD-aware detectors. Furthermore, we investigate a variety of strategies for generating additional training data - from synthetic data injections to autolabelling techniques - to improve the model's ability to recognize and handle OOD objects.

We summarize our contributions as follows:
\begin{itemize}
    \item Designing a new method for out-of-distribution 3D object detection.
    \item Setting a rigorous evaluation protocol by examining evaluation hyperparameters.
    \item Evaluating generation methods of data for unsupervised training an OOD-aware 3D object detector.
\end{itemize}

\section{Related Works}
\subsection{3D Object Detection}
Recent advancements in 3D object detection have led to the development of several notable architectures that enhance the accuracy and efficiency of LiDAR-based perception systems. VoxelNet \cite{Zhou_Tuzel_2018} introduced the concept of end-to-end learning directly from raw point cloud data by voxelizing the space, which significantly improved feature extraction capabilities. Building on this, PointPillars \cite{Lang_2019_CVPR} streamlined the voxelization process by converting point clouds into pseudo-images, enabling faster and more efficient processing suitable for real-time applications. CenterPoint \cite{Yin_2021_CVPR} further improved detection accuracy by proposing a center-based representation, simplifying object localization. Additionally, DSVT~\cite{wang2023dsvt} 
introduced a transformer-based approach to dynamically process sparse voxel grids, enhancing the detection system's robustness and scalability.

\subsection{Open-Set and Out-Of-Distribution Detection}
Several works have advanced open-set detection in 2D out-of-distribution (OOD) detection. VOS \cite{du2022vos} introduced a technique to synthesize virtual outliers, improving a model’s capability to distinguish between known and unknown classes by exposing it to artificially generated outlier examples during training. However, the method contains many hyperparameters for defining class-conditional Gaussian distributions, and they find that fixed Gaussian noise achieves close results. ODIN \cite{odin} enhanced confidence estimates for OOD detection using temperature scaling and input perturbation, leading to more reliable detection of OOD samples. MSP~\cite{msp} proposed using the maximum softmax probability as a baseline for OOD detection. \etal{Liu} \cite{energy} leveraged neural network output energy scores to differentiate between in-distribution and OOD samples. \cite{hendrycks17baseline} provided a robust baseline by analyzing softmax scores to detect misclassified and OOD examples. On the other hand, ODIN, MSP, and Energy thresholding are based on in-distribution data and lack any integration of the \textit{semantic} information of unknown objects. Finally, \cite{towardsopen} addressed the broader challenge of dynamically recognizing new, unseen classes, proposing methodologies to handle an evolving set of object categories in open-world scenarios.

In 3D object detection, handling OOD scenarios has also seen substantial progress. \etal{Cen} \cite{openset3Dobj} explored methods for detecting and managing unknown objects within 3D environments. However, their evaluations are based on UDI, a private dataset with only 600 unknown test objects, and KITTI dataset~\cite{Geiger2012CVPR}. UFO \cite{UFO} introduced novel approaches for identifying unknown objects in point clouds, focusing on accurately segmenting foreground objects, but they also focus only on the KITTI dataset~\cite{Geiger2012CVPR}. \etal{Huang} \cite{Huang2022OutofDistributionDF} explored different instance injection methods to test OOD baselines. The variance in their results underlines the difficulty in identifying the OOD objects they insert from different sources, which have different representations from the KITTI dataset in which they are injected. Lastly, \etal{Kösel} \cite{revisiting} offered an evaluation of existing OOD techniques in the context of LiDAR data, identifying key challenges and proposing enhancements. However, their work lacks specifications on their evaluation protocol and evaluated objects, such as the recall/hit percentage of unknowns, and simply resizes known objects to generate unknown ones and evaluates their method exclusively on nuScenes, which has certain limitations, as we will discuss in Section~\ref{sec:limited_datasets}. We aim to improve the evaluation methodology behind their work as well as the performance through a rigorous benchmark of unknown generation. We find that using a focal loss instead of a binary cross-entropy loss yields better results. Furthermore, their resizing unknown generation method produces lesser results than top-K labelling, suggesting that using pseudo-labels from real OOD data can lead to better performance than reshaping known objects.

\begin{figure*}[!ht]
    \centering
    \includegraphics[width=1\textwidth]{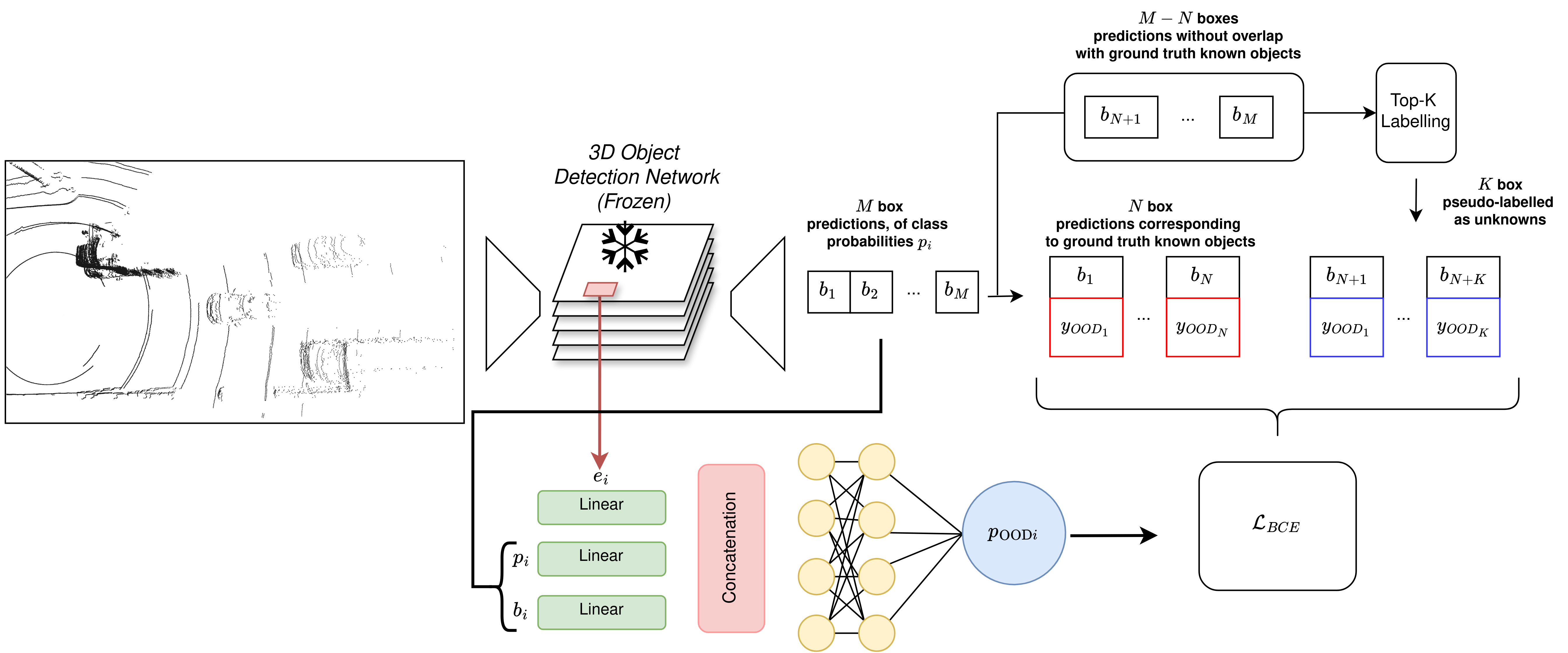}
    \caption{The training pipeline of HD-OOD3D. $p_i$ represents class logits, $e_i$ the embedding of the object extracted from the initial predictions, $b_i$ the predicted box dimensions, and $y_{OOD_i}$ the binary class representing known and unknown objects.}
    \label{fig:training}
\end{figure*}

\section{A new method for 3D OOD Detection}\label{sec:comparison}

While traditional open-set detectors add an extra “unknown” class, we instead view the challenge as an OOD prediction problem. Closed-set detectors are optimized for a fixed set of “core” classes, but an OOD framework allows the model to flag deviations from the learned distribution without sacrificing known-class performance. Methods to compute "unknown" object scores can be generally separated into two categories: single-stage methods, which directly use the outputs of a detector model and apply some thresholding or entropy computation, and two-stage methods, which add additional trainable parameters on top of the base detector.  This latter approach necessitates a retraining phase.

In this section, we introduce our new approach for OOD detection and compare methods that compute “unknown” object scores directly from the detector’s outputs (single-stage methods) with those that add an extra trainable module (two-stage methods).


\subsection{Our approach}\label{sec:our_approach}
We introduce our proposed method \textit{High-Dimensional OOD 3D Object Detection} (HD-OOD3D), which integrates an additional Multi-Layer Perceptron (MLP) module after the base detector. This module is specifically trained to output a binary classification confidence score or an "unknown" confidence score: 0 for known objects and 1 for unknown objects. We illustrate the training pipeline of HD-OOD3D in Figure~\ref{fig:training}. 
This two-stage solution employs the MLP to process initial detections, decomposed into embeddings $e_i$ extracted at the feature map locations of predicted bounding boxes, box predictions $b_i$, and prediction logits $p_i$ and outputs an "out-of-distribution" score $p_\text{OOD}$. During inference, if $p_\text{OOD}$ is larger than a given threshold $\text{OOD}_\text{thresh}$, the object is considered an OOD object.

The closest approach to our method is MMOOD3D~\cite{revisiting}, which is also a two-stage method. The main differences with MMOOD3D are:
\begin{itemize}
    \item high-dimensional feature map (\textit{High-Dim}): we use higher-dimension (512) feature map representation to extract the embeddings $e_i$ from the base detector instead of the final lower-dimension (192) feature maps 
    \item feature interpolation (\textit{Interp.}): instead of extracting the feature representation of the base detector's boxes at the discrete grid value of the feature map, we interpolate these features based on the predicted center
    \item a focal loss instead of a binary cross-entropy loss (\textit{F. Loss})
    \item a $3\times3$ max-pooling (\textit{Pool}) of the feature map before extracting the feature vector at the corresponding feature map location of the predicted object center
\end{itemize}
Section~\ref{subsec:input_choices} discusses the contribution of these modifications.
Section~\ref{subsec:res} also presents a comparison between our approach to unsupervised learning of unknowns using top-K labeling and the MMOOD3D method~\cite{revisiting}, which constructs unknown objects by resizing known ones.

\subsection{Evaluation with supervised unknown objects}

First, using a methodology we will refer to as "supervised," we will leverage the labels of the unknown objects in the chosen dataset. This supervised approach allows for a more rigorous comparison of the different methods: one-stage, two-stage, and our HD-OOD3D method. In the second part, Section~\ref{sec:gendata}, we will explore the "unsupervised" methodology, where we assume we do not have access to the labels of the unknown objects in the dataset. This will enable us to compare various pseudo-label generation methods for unknown objects.


\subsubsection{The limits of widely used datasets}\label{sec:limited_datasets}
Currently, when addressing the out-of-distribution (OOD) detection problem in 3D object detection, the limitations of existing datasets such as KITTI \cite{Geiger2012CVPR}, ONCE \cite{once_ds}, and Waymo \cite{Sun_2020_CVPR} become particularly pronounced. Indeed, these datasets contain a small number of classes, respectively 5, 5, and 3. Methods such as \cite{openset3Dobj}, which evaluate on the \textit{Van} and \textit{Truck} subset of KITTI, tackle a specific part of the issue, these classes being both vehicles and lacking the high diversity which can arise in real-world scenarios. 

nuScenes~\cite{caesar2020nuscenes} is one of the most used datasets in autonomous driving but presents specific challenges critical in the context of OOD detection. Firstly, the predefined dataset splits and evaluation metrics in nuScenes might not be sufficient for evaluating OOD performance. 
For instance, only one out of four of the dataset's scans in its evaluation set contains unknown objects as defined by \cite{revisiting}, thus limiting the evaluation interpretation. nuScenes also suffers from its point cloud sparsity due to its 32-beam LiDAR sensor, which can only be solved by aggregating scans and creating a trailing point cloud for moving objects, hindering detection.
These findings underscore the need to use a more diverse and dense dataset to effectively tackle the OOD problem in 3D object detection.

\subsubsection{Argoverse 2 dataset for OOD evaluation}

In order to establish a robust baseline for out-of-distribution (OOD) detection, we turn to the Argoverse 2 dataset \cite{wilson2021argoverse}, which offers several distinct advantages over other datasets such as nuScenes. Argoverse 2 is noted for its high class diversity, encompassing a wide range of object types and scenarios, which is crucial for training models that can generalize well to new, unseen objects. This dataset also provides better LiDAR representation, with higher density point clouds than nuScenes, due to its use of two 32-beam LiDAR sensors, without the "point cloud trail" phenomenon. The dataset contains $110,071$ scans in the training set and $23,547$ in its validation set. We underline its appeal compared to nuScenes in Table \ref{tab:hits}, which shows the percentage of ground truth objects used for metric computation as per the matching and evaluation procedure described in \ref{subsec:matching}, i.e. detection hits. The percentage of total ground truth unknown objects that will be considered for the evaluation is twice as low for nuScenes compared to Argoverse 2, showing more limited interpretability of results.

\begin{table}[ht]
\centering
\scalebox{1}{
\begin{tabular}{l |  c  c c  }
Dataset & Hits \% & Unknown objects & Known objects \\
\hline
\multirow{1}{*}{Argoverse2} 
& & 24.1&50.2\\
\hline
\multirow{1}{*}{nuScenes} 
&  & 12.9&45.0\\

\hline
\end{tabular}
}
\caption{A comparison between the percentage of detection hits for Argoverse 2 and nuScenes in our evaluation procedure, using the outputs of an initial prediction model CenterPoint along with our final evaluation hyperparameters}
\label{tab:hits}
\end{table}

For Argoverse 2, we choose the rarest 5\% of classes as our unknown classes, denoted as set $U$. This approach ensures that the chosen unknown classes are genuinely underrepresented in the training data, providing a challenging test for the OOD detection capabilities of our models. This selection method helps simulate real-world scenarios where we assume that the most common objects will have been labeled beforehand while certain rare objects are infrequently encountered, making it a valuable benchmark for assessing the robustness of 3D object detection systems. The unknown classes $U$ (11 labels), representing $28,271$ instances in the validation set, are:

\noindent\begin{minipage}{\linewidth}
\footnotesize
\begin{align*}
U &= \{ \text{Motorcyclist, School bus, Message board trailer, Truck cab,}\\
&\quad \text{Articulated bus, Stroller, Motorcycle, Mobile pedestrian crossing }\\
&\quad \text{sign, Wheeled rider, Wheelchair, Dog} \}
\end{align*}
\end{minipage} \\

The closed "known" classes, denoted as set $C$ (15 labels), include more common objects, presumed to have been labeled beforehand. These classes, representing $1,391,546$ instances in the validation set, are:

\noindent\begin{minipage}{\linewidth}
\footnotesize
\begin{align*}
C &= \{ \text{Regular vehicle, Pedestrian, Bollard, Construction cone,}\\
&\quad \text{Stop sign, Sign, Bus, Truck, Bicycle, Bicyclist, Wheeled device,}\\
&\quad \text{Box truck, Large vehicle, Construction barrel, Vehicular trailer} \}
\end{align*}
\end{minipage} \\

\subsubsection{Implementation and metrics}

In the first stage, we train a CenterPoint~\cite{Yin_2021_CVPR} object detector on the Argoverse 2 dataset~\cite{wilson2021argoverse} with classical training regimes~\cite{revisiting, Yin_2021_CVPR}. Our implementation is based on the OpenPCDet framework~\cite{openpcdet2020}. Specifically, CenterPoint is trained and evaluated in the $\left[-72.0, 72.0\right]  \mathrm{m}$ x-y range and uses a voxel size of $\left[0.1, 0.1, 0.2\right] \mathrm{m}$. We augment the data using random flipping along the x-axis, rotation around the z-axis, and random scaling of the point clouds in the $\left[0.95, 1.05\right]$ range. We use the Adam optimizer with a OneCycle learning rate, initialized at $1\mathrm{e}{-3}$, a weight decay of $0.01$, momentum of $0.9$, and train for ten epochs, using a focal loss to compute the gradients.

This model obtained a mAP score of $35.2$ on the known classes (15 labels). We freeze this first stage model then add a 3-layer MLP where we divide the dimensionality by two between every layer ($512 \rightarrow 256 \rightarrow 128 \rightarrow 2$), followed by a sigmoid activation function. We train it for five epochs, with a polynomial learning rate using a power of $3$, with an initial learning rate of $1\mathrm{e}{-3}$.
 
The proposed second-stage neural network adds only about 500k parameters and a few milliseconds of inference time, both negligible compared to the 30M parameters and latency (around 50~ms) of the base CenterPoint detector. This demonstrates that two-stage methods can provide OOD capability with minimal overhead, making HD-OOD3D feasible for real-world deployment in autonomous vehicles.

We compared multiple single-stage methods for ODD detection and also the two-stage method MOOD3D~\cite{revisiting} against our approach on the Argoverse 2 dataset with our evaluation protocol of Section~\ref{sec:exp_gt}. We selected several single-stage methods to ensure a comprehensive assessment, i.e. methods that directly use the output logits of the detection model. The chosen methods include ODIN \cite{odin}, MaxLogit \cite{msp}, Energy Thresholding \cite{energy}, MC Dropout inspired by \cite{mc_dropout}, and the Default Score.

We assess the OOD detector using several metrics: AUROC, which quantifies the overall likelihood that a positive case is ranked above a negative case; FPR-95, which indicates the false positive rate when the true positive rate is set at 95\%, balancing sensitivity and specificity; and the AUPR metrics, AUPR-E and AUPR-S, which focus on the model’s precision-recall balance for unknown and known objects respectively. Importantly, AUPR-S represents the model performance on known objects. If a noticeable downgrade in this metric is noticed, such as under 99, then we consider the base detector's performance on known objects to be hindered by the additional unknown object classifier module. The threshold $\text{OOD}_\text{thresh}$ is dynamically tuned at evaluation to compute metrics like AUROC, FPR-95, and AUPR.

\subsection{Results}\label{sec:single_stage}

Table \ref{tab:baseline_res} shows the results. All single-stage models fail to reach the performance of two-stage methods. While ODIN achieves the highest AUROC among single-stage approaches (85.9 on Argoverse 2), it still exhibits a high False Positive Rate at 95\% recall (61.6\%). Energy-based methods, Monte-Carlo Dropout, MaxLogit, and Default Score thresholding yield relatively weaker results, with AUROC values ranging from 65.8 to 76.6. In contrast, two-stage methods significantly outperform single-stage approaches. MMOOD3D reaches an AUROC of 88.6 with a lower FPR-95 of 43.2. Our proposed method, HD-OOD3D, achieves competitive results with an AUROC of 86.8 while maintaining a lower FPR-95 (51.1) compared to ODIN. Crucially, HD-OOD3D preserves the performance on known objects (AUPR-S = 99.7), whereas MMOOD3D suffers a substantial drop in AUPR-S to 70.5. This suggests that HD-OOD3D provides a better balance between detecting unknown objects and preserving the accuracy of the base detector.

\begin{table}[ht]
    \centering
    \scalebox{0.78}{
\begin{tabular}{l |l|c c c c c}
Type & Method & AUROC $\uparrow$ & FPR-95 $\downarrow$&  AUPR-E $\uparrow$ & AUPR-S  $\uparrow$ \\
\hline
 & Default Score & 76.6&86.5&4.3&99.4\\
Single & Energy \cite{energy}& 65.8 & 94.6 & 2.7 & 99.0\\
Stage & Monte Carlo Dropout \cite{mc_dropout} & 75.4& 69.7& 7.0& \cellcolor{yellow!40}98.9\\
& MaxLogit \cite{msp}& 76.6 & 86.5&4.3& \cellcolor{yellow!40}98.6\\
& ODIN \cite{odin}& 85.9&61.6& 9.8 & \cellcolor{yellow!40}97.1\\

\hline
Two& MMOOD3D \cite{revisiting} & 88.6 &43.2 &27.1& \cellcolor{yellow!40}70.5 \\
Stage& \cellcolor{gray!20}HD-OOD3D (Ours)& \cellcolor{gray!20}86.8& \cellcolor{gray!20}51.1& \cellcolor{gray!20}20.0& \cellcolor{gray!20}99.7\\
\end{tabular}}
    \caption{
    Comparison of AUROC, FPR-95, AUPR-S, AUPR-E metrics for single-stage and two-stage methods on Argoverse 2 in the supervised setup, where ground truth labels for unknown objects are used during training. The yellow cells indicate methods with an AUPR-S score below 99, which means that the base detector has lost performance on the known classes.}
    \label{tab:baseline_res}
\end{table}


\subsection{Ablation study on HD-OOD3D}\label{subsec:input_choices}

In Table~\ref{tab:model_param}, we compare different parts of our approach described in Section~\ref{sec:our_approach}.

\begin{table}[ht]
    \vspace{0.3cm}
    \scalebox{0.77}{
\begin{tabular}{c c c c c c c c c}
High-Dim & Interp. & F. Loss & Pool & AUROC $\uparrow$ & FPR-95 $\downarrow$&  AUPR-E $\uparrow$ & AUPR-S  $\uparrow$ \\
\hline
&  &  & & 85.2& 52.6& 17.6& \cellcolor{yellow!40}70.4\\
\checkmark&  &  & & 77.3& 66.9& 10.2& \cellcolor{yellow!40}79.1\\
&  \checkmark&  & & 85.4&51.6&17.8&\cellcolor{yellow!40}70.6\\
&  & \checkmark & & 86.8& 50.6& 15.7& 99.7\\
&  &  & \checkmark& 88.7& 44.5& 26.5&\cellcolor{yellow!40}71.8\\
 \hspace{1.5mm}\checkmark$^*$ &\hspace{1.5mm}\checkmark$^*$ &  & & 88.6&43.2&27.1&\cellcolor{yellow!40}70.5\\
  \cellcolor{gray!20}\checkmark &  \cellcolor{gray!20}\checkmark &  \cellcolor{gray!20}\checkmark &  \cellcolor{gray!20}\checkmark & 88.6 & 51.1 & 20.0 & 99.7 \\
\end{tabular}}
    \caption{Comparison of AUROC, FPR-95, AUPR-S, AUPR-E metrics for different variations of an MLP two-stage method on Argoverse 2. The $^*$ configuration represents method MMOOD3D~\cite{revisiting} while the grey line is our approach HD-OOD3D. The yellow cells indicate methods with an AUPR-S score below 99, which means that the base detector has lost performance on the known classes.
}
    \label{tab:model_param}
\end{table}


The results show that adding a 3x3 pooling layer and a focal loss was beneficial, compared to the work~\cite{revisiting}, which only uses a binary cross-entropy loss, no pooling, high-dimensional feature maps, and feature interpolation. 
Despite a higher FPR-95, this is the only method that combines a higher precision on known objects and near-perfect precision on closed objects. 
For instance, in the Argoverse 2 dataset, a $3\times3$ maxpooling filter enabled a feature aggregation operation with a radius of 2.1 meters. It substantially improved OOD performance by ensuring a more robust aggregation of surrounding point cloud data.  


\section{Analysis of the Evaluation Protocol for 3D OOD}\label{sec:exp_gt}

Here, we examine how external factors and hyperparameter choices influence OOD detection performance because these elements directly affect both the quality and reliability of the detection metrics. For instance, the characteristics of the dataset, such as point cloud density, class diversity, and the presence of noise, can alter how many true and false positives are identified. Similarly, hyperparameters like the distance threshold for matching predictions, the detection score cutoff, and the criteria used for sorting predictions determine which detections are considered valid. Even small adjustments in these parameters can lead to significant shifts in performance metrics, affecting the reproducibility and fairness of the evaluation. Therefore, we examine and compare these choices.

\subsection{Metrics and Matching}\label{subsec:matching}
As previously described, we use the AUROC, FPR-95, AUPR-E (with
unknown classes as positives) and AUPR-S (with known classes
as positives) to assess the model's ability to distinguish between known and unknown objects. We also describe our matching algorithm, detailed in Algorithm \ref{ref:algo}, which is crucial for computing True Positives (TP), False Positives (FP), True Negatives (TN), and False Negatives (FN). This algorithm, which includes essential hyperparameters, plays a significant role in determining the final evaluation metrics, mainly by ignoring any predicted box whose center is not within a fixed range of any non-assigned ground truth.

\begin{algorithm}[t]
\caption{Matching and Evaluation Algorithm}
\small 
\ForEach{point cloud $\mathcal{P}$ and its associated ground truth (GT) bounding boxes in the chosen validation set $\mathcal{V}$}{
    Compute Euclidean distance between the M predictions (objects whose base detector prediction score exceeds $\delta_{\text{thresh}}$) and the $N$ GTs. 
    Sort by order of confidence the distances, the boxes, and "unknown" scores $p_\text{OOD}$\;
    \ForEach{sorted prediction pred}{
        \eIf{all GTs already matched with pred}{
            End\;
        }{
            \eIf{dist(pred, closest unmatched GT)$<$ $d_\text{thresh}$}{
                \eIf{$p_\text{OOD} > \text{OOD}_\text{thresh}$}{
                    pred unknown\;
                }{
                    pred known\;
                }
                pred known, GT unknown $\rightarrow$ TP \\
                pred known, GT unknown $\rightarrow$ FP \\
                pred unknown, GT unknown $\rightarrow$ TN \\
                pred unknown, GT known $\rightarrow$ FN \\
            }{Continue\;}
        }
    }
}
\normalsize 
\label{ref:algo}
\end{algorithm}


We use the ground truth data to train the MLP with the actual annotated 'unknown' classes, like in Section \ref{sec:comparison}.
We demonstrate that many external factors could impact the performance and resulting metrics, which are not disclosed in \cite{revisiting}. We identify several key parameters that significantly influence the effectiveness of HD-OOD3D when applied to GT data, allowing us to examine the influence of these parameters independently of the actual performance of the MLP. 

 For the two-stage family of methods, we train an MLP after freezing the initial detector Centerpoint, like in the previous section, and use the same MLP architecture. For ease of reading, we include the relevant experiments from our two-stage HD-OOD3D in the following.

\subsection{Hyperparameter Analysis}
By systematically analyzing the evaluation hyperparameters, we aim to provide a more comprehensive understanding of how different evaluation factors influence the final performance of unknown OOD detection. These hyperparameters include:
\subsubsection{Evaluation set}

Using the "Open" Scans subset, which includes only scans with unknown objects, significantly improved the AUPR-E metric from 12.6 to 20.0 for HD-OOD3D. Due to the known/unknown class imbalance, using all scans adds many more potential false positives without any unknown ground truth, hindering all methods across the board. Consequently, our final evaluation used the "open" subset, only the evaluation scans containing at least one unknown object, to better assess performance. It is important to note that, as shown in Table \ref{tab:fused_tab}, the chosen hyperparameters do not maximize the performance of HD-OOD3D in isolation, but rather provide a balanced and fair benchmark for all methods.

\subsubsection{Distance and Threshold Settings for OOD Detection}
The choices of detection module threshold $\delta_{\text{thresh}}$ (threshold above which the boxes are considered as objects by the base detector) and distance metric $d_\text{thresh}$ (distance used for associating predicted boxes with ground-truth boxes) play a critical role in OOD Detection.

Adjusting the distance threshold for matching predicted boxes to ground truth objects influences the inclusion of predictions in metric calculations. Specifically, when using a larger $d_\text{thresh}$ (e.g., 2.0m instead of 0.5m), more predictions, mainly those correctly identifying unknown classes, are included. This increase in detection hits, predicted boxes within the range of ground truth, allows for a more comprehensive evaluation of the model's ability to detect unknown objects. However, this also introduces a risk of misassigning predictions to known objects, potentially lowering AUROC and FPR-95 while improving AUPR-E, as this metric directly reflects the precision in detecting unknowns.

A lower threshold $\delta_{\text{thresh}}$ in the detection module (e.g., 0.3) allows more predicted objects to pass through, reducing misses, especially for known objects. However, this comes at the cost of an increased likelihood of false positives among the unknown detections. A trade-off exists: while a lower threshold increases the detection rate of unknown objects, it also results in more potential false positives. Therefore, a threshold of 0.3, paired with a 2.0m distance for box matching, strikes a balance between minimizing misses for known objects and enhancing the detection of unknown objects, as evidenced by the metrics in Table \ref{tab:fused_tab}. 

\begin{table}[ht]
\centering
\scalebox{0.7}{
\begin{tabular}{l | l | c  c | c c c c }
$\delta_{\text{thresh}}$ & $d_\text{thresh}$& Hits (U)\%& Hits (C)\%  & AUROC $\uparrow$ & FPR-95 $\downarrow$&  AUPR-E $\uparrow$ & AUPR-S  $\uparrow$\\
\hline
\multirow{2}{*}{0.05} & \multirow{1}{*}{0.5m}& \multirow{1}{*}{22.8}& \multirow{1}{*}{49.3}&\cellcolor{white!20}91.0& \cellcolor{white!20}43.7& \cellcolor{white!20}22.2&\cellcolor{white!20}99.8\\
& \multirow{1}{*}{2.0m}& \multirow{1}{*}{\textbf{38.9}}& \multirow{1}{*}{\textbf{55.4}} & \cellcolor{white!20}84.9&\cellcolor{white!20}62.8& \cellcolor{white!20}17.5&\cellcolor{white!20}99.5\\
\hline
\multirow{2}{*}{\underline{0.3}} & \multirow{1}{*}{0.5m}& \multirow{1}{*}{14.4}& \multirow{1}{*}{45.3} & \cellcolor{white!20}92.3& \cellcolor{white!20}37.4& \cellcolor{white!20}23.2&\cellcolor{white!20}\textbf{99.9}\\
& \multirow{1}{*}{\underline{2.0m}}& \multirow{1}{*}{26.2}& \multirow{1}{*}{50.6} & \cellcolor{white!20}88.6& \cellcolor{white!20}51.1& \cellcolor{white!20}20.0&\cellcolor{white!20}99.7\\
\hline
\multirow{2}{*}{0.5} & \multirow{1}{*}{0.5m}& \multirow{1}{*}{7.6}& \multirow{1}{*}{39.0} & \cellcolor{white!20}\textbf{93.6}& \cellcolor{white!20}\textbf{32.6}& \cellcolor{white!20}\textbf{24.8}&\cellcolor{white!20}\textbf{99.9}\\
& \multirow{1}{*}{2.0m}& \multirow{1}{*}{11.7}& \multirow{1}{*}{42.4} & \cellcolor{white!20}91.4 & \cellcolor{white!20}38.7& \cellcolor{white!20}22.1&\cellcolor{white!20}99.9\\
\end{tabular}
}
\caption{Impact of thresholds $\delta_{\text{thresh}}$ and $d_\text{thresh}$ on Detection Hits (for unknown $U$ instances and closed known $C$ instances), and OOD Metrics for Unknown and Known Ground Truth Objects in the Argoverse 2 dataset. Underlined parameters represent our final hyperparameter choices.}
\label{tab:fused_tab}
\end{table}
\subsubsection{Sorting method}

In our experiments, we observed a 2.0 increase in the AUROC metric when sorting predictions by the base detector's confidence score rather than by the "unknown" score confidence. This suggests that using the original detector's score is more reliable for prioritizing predictions, especially given the high prediction scores for unknown objects and the non-overlap hypothesis in 3D space.

From Table \ref{tab:fused_tab}, we notice widely different results from the traditional OOD metrics for any fixed method, depending on the evaluation hyperparameters. The choice of detection score threshold $\delta_{\text{thresh}}$ and distance threshold $d_\text{thresh}$ is crucial for balancing the precision and recall of OOD detection. Both $\delta_{\text{thresh}}$ and $d_\text{thresh}$ determine how detections are matched to ther ground truth: larger threshold includes more detections but risk misassigning predictions. Rather than focusing on optimizing the performance of our method, our primary goal was to refine the evaluation protocol to ensure a fair and robust comparison. Through systematic evaluation, we found that setting $\delta_{\text{thresh}} = 0.3$ and $d_\text{thresh} = 2.0$m provides the best trade-off, maximizing recall for OOD objects while minimizing degradation on known classes. These choices allow for a more reliable assessment of OOD detection performance across different methods. We suggest that future works consider the recall of the base detector and explicitly specify these crucial evaluation hyperparameters to facilitate reproducibility and fair benchmarking.

\section{Comparison of Methods for Unsupervised Learning of Unknown 3D Objects}\label{sec:gendata}

In the previous sections, our MLP models were trained using ground truth labels for unknown objects. However, in real-world scenarios, manually annotating every unknown object is infeasible due to the diversity and unpredictability of OOD instances. Unlike in controlled datasets like Argoverse 2, where unknown objects can be labeled explicitly, a practical deployment scenario requires detecting OOD objects without prior supervision. To address this challenge, we propose generating pseudo-labels for unknown objects, enabling training without explicit ground truth annotations.

\subsection{Different Representations of Out-of-Distribution Objects}

 
\begin{figure*}[ht]
    \begin{center}
        \hspace{0cm}
    \begin{subfigure}{0.31\textwidth}
        \includegraphics[width=\textwidth]{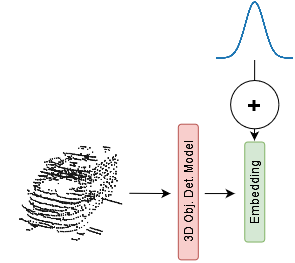}
        \caption{Noise in latent space}
    \end{subfigure}
    \hspace{1.5cm}
    \begin{subfigure}{0.31\textwidth}
        \includegraphics[width=\textwidth]{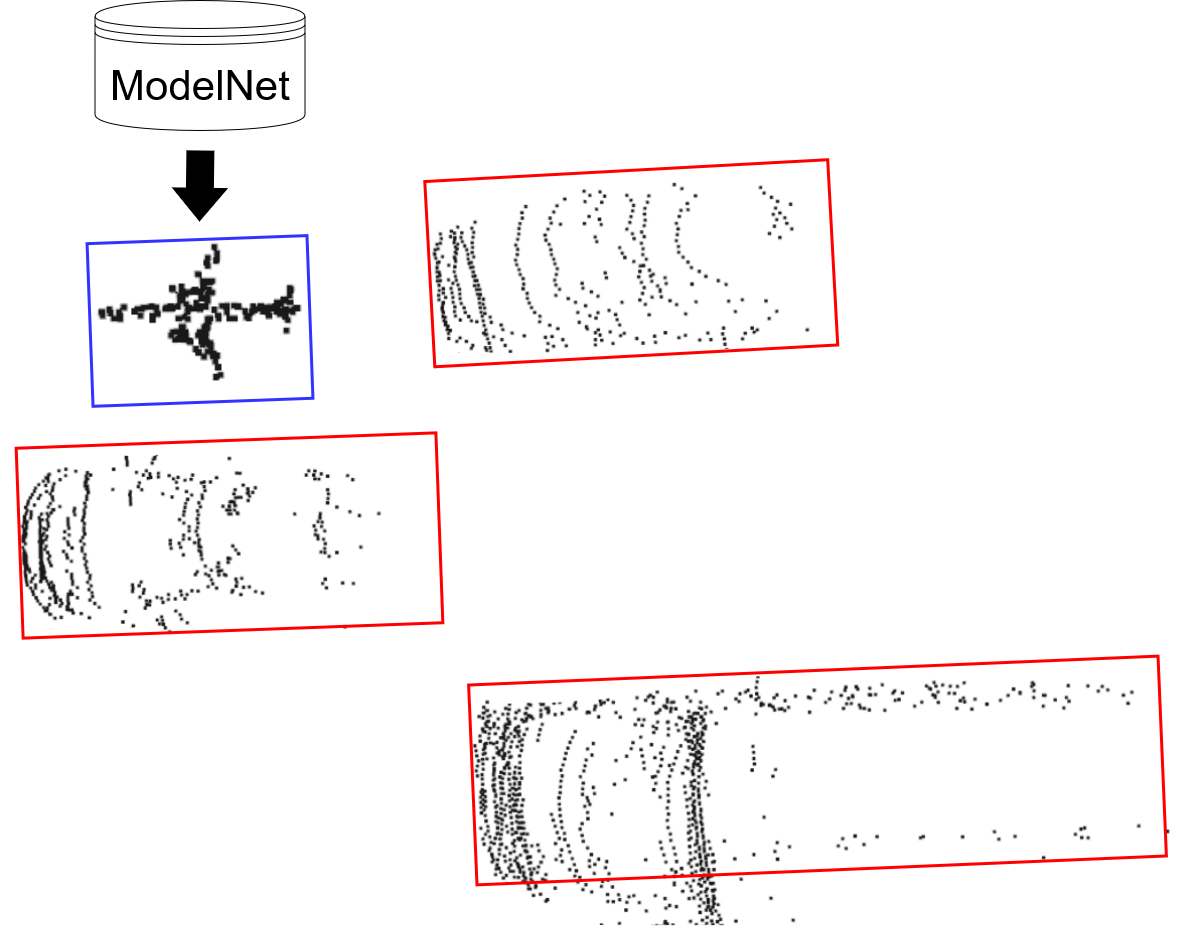}
        \caption{External Data from ModelNet~\cite{modelnet}}
    \end{subfigure}
    \end{center}

    ~
    \begin{center}
        \begin{subfigure}{0.31\textwidth}
        \includegraphics[width=\textwidth]{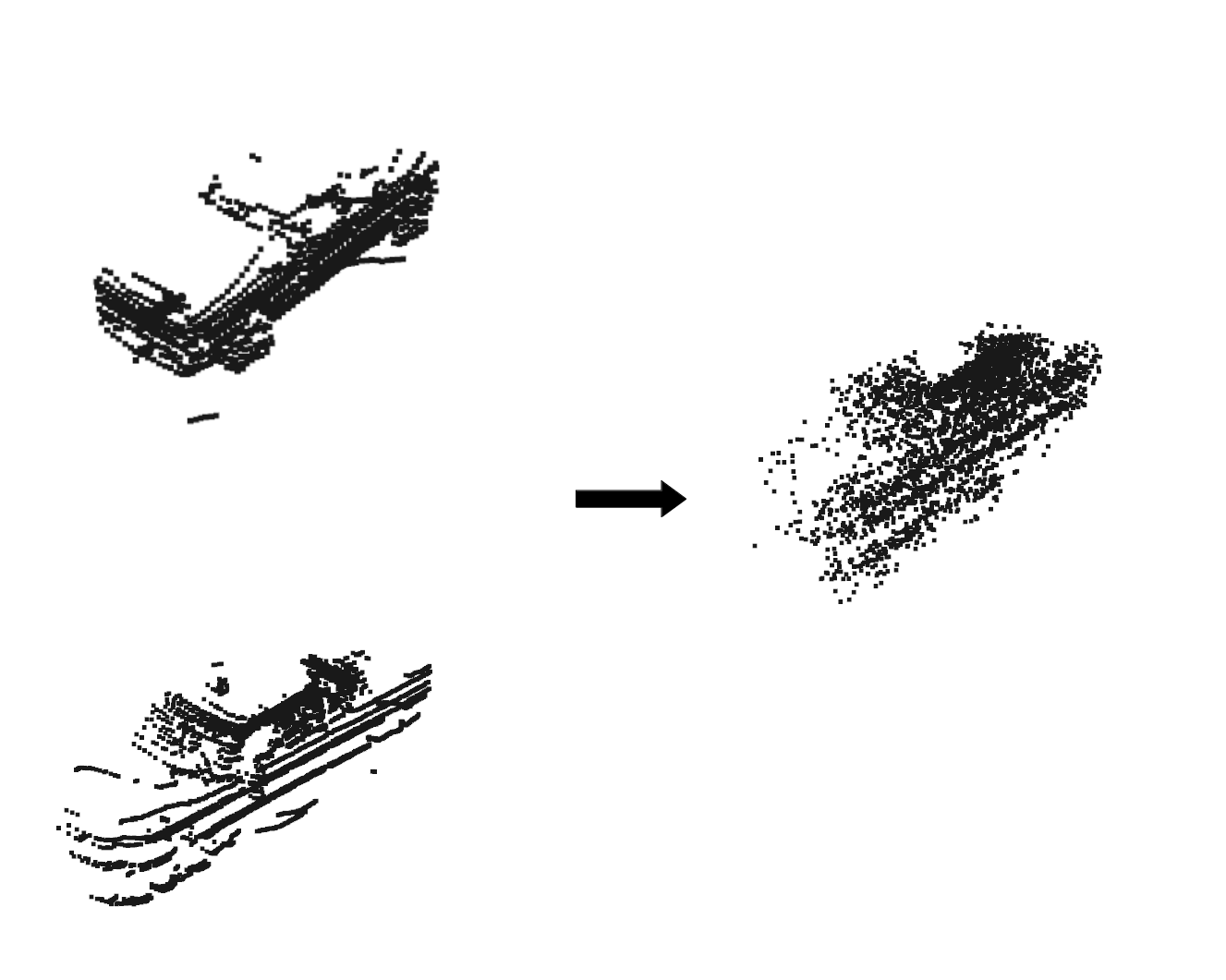}
        \caption{PointMixup}
    \end{subfigure}
    \hfill
    \begin{subfigure}{0.3\textwidth}
        \includegraphics[width=\textwidth]{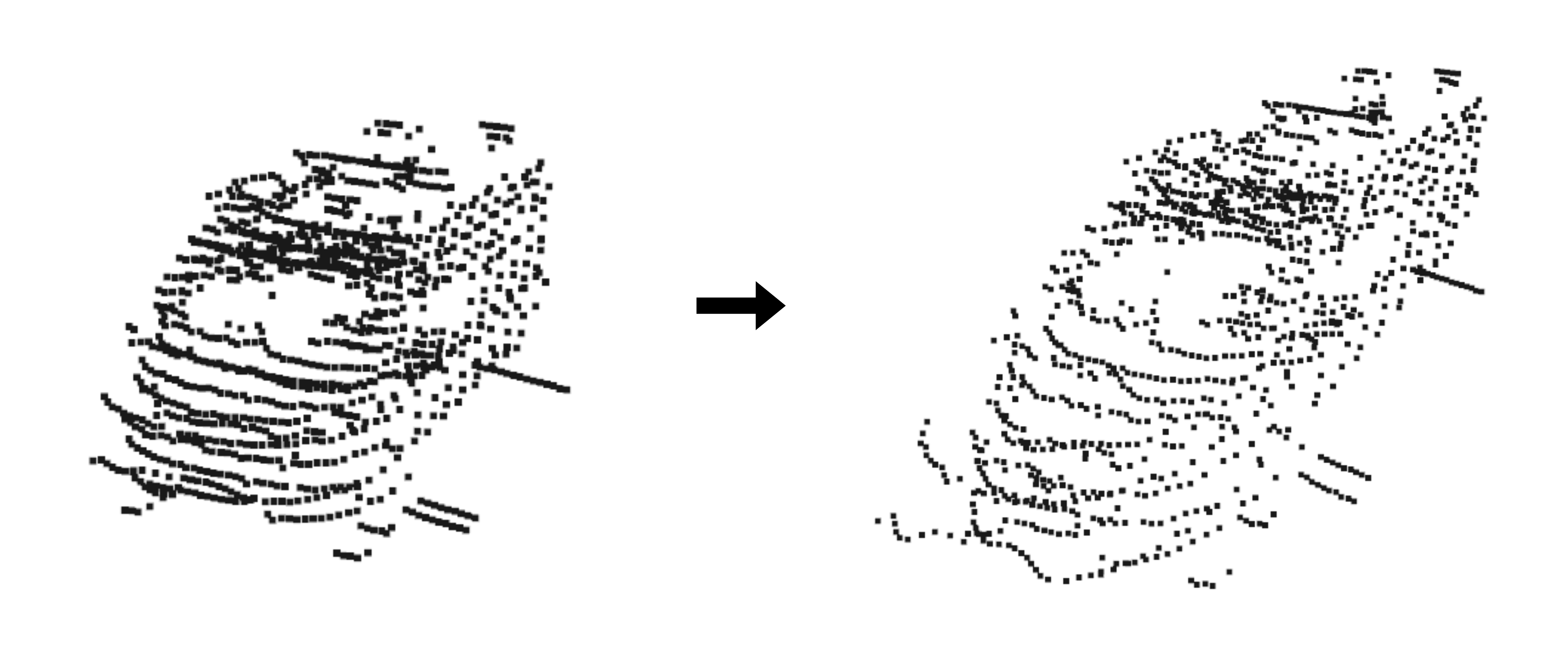}
        \caption{Resizing}
    \end{subfigure}
    \hfill
    \begin{subfigure}{0.36\textwidth}
        \includegraphics[width=\textwidth]{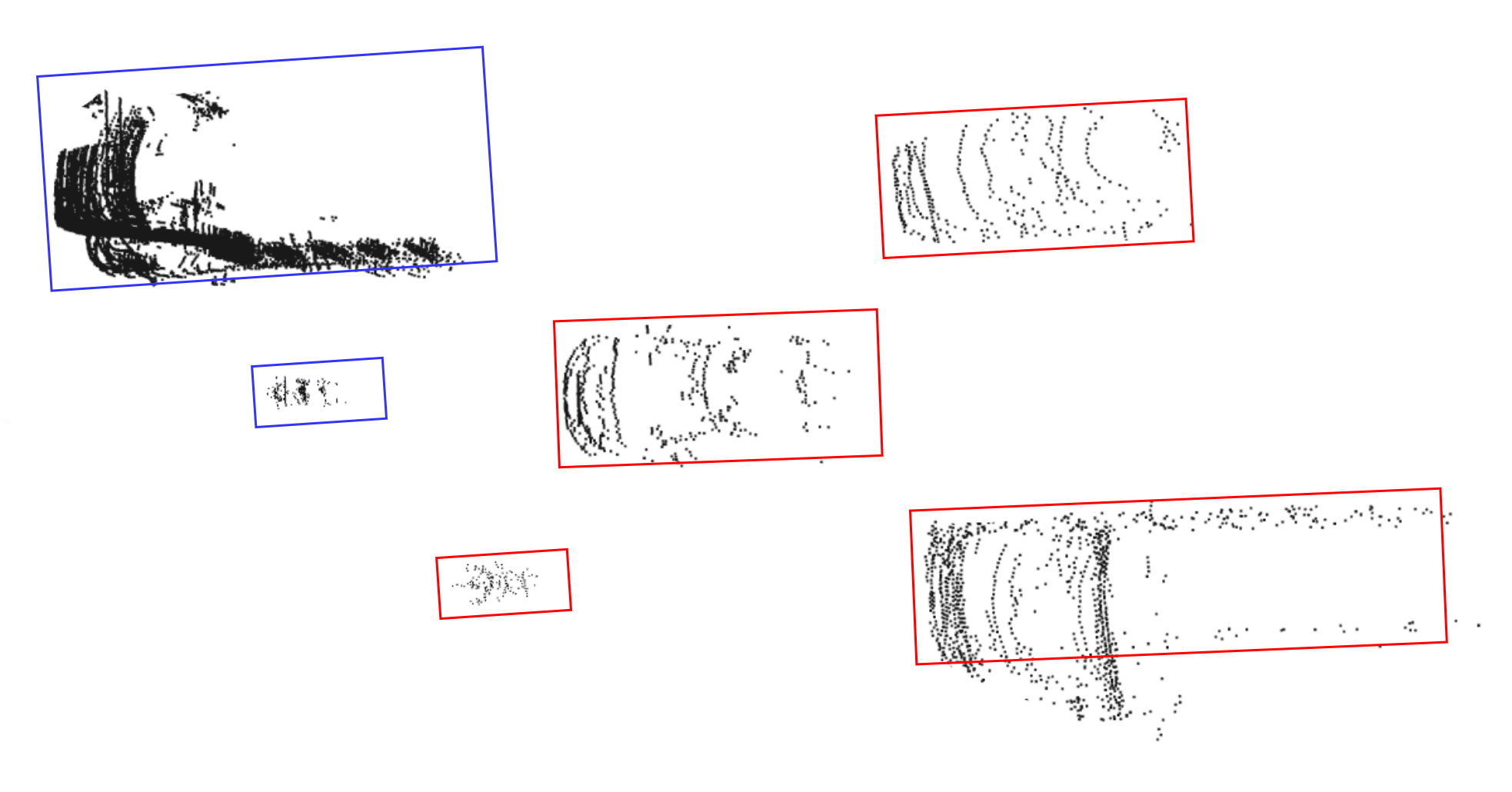}
        \caption{Top-K Autolabelling}
    \end{subfigure}
    \end{center}
    
    \caption{Generation Methods for Unknown Objects in 3D LiDAR to Supervise Two-Stage OOD Detection}
    \label{fig:generation}
\end{figure*}

In contrast to the single-generation approach used in MMOOD3D \cite{revisiting}, we propose several alternative methods to synthesize out-of-distribution (OOD) objects.
We compare these different representations of out-of-distribution objects for training, which we create through several means:
\begin{itemize}
    \item Gaussian noise \cite{du2022vos}: we perturb the extracted embedding of ID objects with Gaussian noise, adding a normal distribution $\mathcal{N}(0, 1)$  to the extracted embeddings. This approach allows us to inject variability into the feature space, thereby creating synthetic representations of unknown objects.
    \item External data: we incorporate ModelNet~\cite{modelnet}, allowing us to introduce a diverse set of synthetic objects into our training regime, which we consider OOD. We insert between 15 and 25 objects, sampled randomly from the \{\textit{'bench'}, \textit{'bookshelf'}, \textit{'chair'},\textit{ 'cone'}, \textit{'desk'}, \textit{'flower pot'}, \textit{'lamp'}, \textit{'mantel'}, \textit{'night stand'}, \textit{ 'plant'},\textit{ 'vase'},\textit{ 'wardrobe'}\} class subset of ModelNet, and normalize their size to 1. We then sample 200 points at the surface of each object, followed by a grid sampling where we keep between 1 and 3 points with a grid size which we also randomize between 5 and 10, apply an intensity value to the points based on the intensity statistics of Argoverse 2, and finally resize the objects with a global scaling factor between 1 and 4.
    \item PointMixup \cite{pointmixup}: we use this method to mix point clouds of different objects through point matching and interpolation to create new, hybrid representations. We expect objects created in such a fashion to contain local features present in both ID and OOD objects while creating novel appearances. Since \cite{pointmixup} was designed for data augmentation, we chose different interpolation hyperparameters after testing variations. Each object with at least 5 points has a 20\%  probability of being replaced by a new version of that object generated as an interpolation between it and another object with at least 5 points, with interpolation factors randomly chosen in the $[0.3, 0.7]$ range. 
    \item Resizing: we create out-of-distribution objects by resizing in-distribution (ID) known objects to modify their scales along all three axes, similarly to \cite{revisiting}. Each known object with at least 5 points has a 50\% probability of being resized, where for each axis, we randomly apply a scaling factor between $[0.1, 0.5]$ and $[1.5, 3.0]$. This simple transformation creates new, out-of-distribution representations by altering the spatial dimensions of familiar objects.
    \item Our 3D Top-K autolabelling : inpired by \cite{towardsopen} for open 2d objet detection in images, we propose in 3D to use the top-K most confident predictions from the base detector, which don't overlap with ground truth known objects, as OOD objects. In this approach, we keep the scans from the datasets containing real OOD distribution, as we expect these to become autolabelled. 
    
\end{itemize}

We illustrate these different generation methods in Figure \ref{fig:generation} and compare these various representations of OOD objects and their impact on the training and performance of our additional module on real validation OOD data. All trainings are done with our approach HD-OOD3D of Section~\ref{sec:our_approach}.

\subsection{Results}\label{subsec:res}

The results of our evaluation on the Argoverse 2 dataset, presented in Table \ref{tab:res_table}, highlight the discrepancies in the MLP performance based on different generation methods of unknowns.

The Resizing method obtained an AUROC of 77.2, with a relatively low AUPR-E of 6.3. These low results highlight how this simple approach might not be suitable for the more diverse Argoverse 2. The Gaussian Noise method performed poorly across all metrics, with an AUROC of 66.5 and a very high FPR-95 of 84.5, making it only slightly better than random chance. 

\begin{table}[ht]
\scalebox{0.8}{
\begin{tabular}{l|c c c c}
Methods & AUROC $\uparrow$ & FPR-95 $\downarrow$&  AUPR-E $\uparrow$ & AUPR-S  $\uparrow$ \\
\hline
Gaussian Noise \cite{du2022vos} & 66.5 & 84.5 & 3.7 & \cellcolor{yellow!40}98.9 \\
ModelNet Injection$^\dag$& 70.6 & 78.7 & 5.0 & \cellcolor{yellow!40}71.1 \\
PointMixUp$^\dag$\cite{pointmixup}&72.1 & 72.4& 7.4 & 99.1\\
Resizing from MMOOD3D\cite{revisiting}& 77.2 & 75.3 & 6.3& 99.4 \\

Top-5 Labelling (Ours) & \textbf{81.9}& \textbf{68.2}& \textbf{8.8}& \textbf{99.6} \\
\hline
GT Unknowns& 86.8& 51.1& 20.0& 99.7\\
\end{tabular}

}
\caption{Comparison of AUROC, FPR-95, AUPR-E, and AUPR-S metrics for different unknown generation methods on Argoverse 2. $^\dag$ methods represent adaptations of data augmentation methods. The yellow cells indicate methods with an AUPR-S score below 99, which means that the base detector has lost performance on the known classes. GT Unknowns uses the labels of the unknown objects to supervise the MLP and can be regarded as an oracle.}
\label{tab:res_table}
\end{table}

ModelNet Injection, which incorporates synthetic objects from the ModelNet dataset as OOD data, showed better performance with an AUROC of 70.6 and an AUPR-E of 5.0. We believe the difficulties in OOD detection for this method stem from the synthetic nature of the dataset, which suggests that while this method might be interesting \cite{UFO}, one has to be careful about the quality of these instances and the domain differences between these and the real unknowns. PointMixUp struggled with a lower AUROC of 72.1 compared to the Resizing method with 77.2, even though it remains the best method for the FPR-95 and AUPR-E metrics compared to previous methods.


The Top-K Labelling method, which labels the top K most confident predictions not overlapping with ground truth known objects as OOD, emerged as a notable performer. It achieved an AUROC of 81.9 and the highest AUPR-S of 99.6, demonstrating good identification of known objects. Despite its higher FPR-95 of 68.2, Top-K Labelling showed promise in balancing the detection of known and unknown objects by leveraging real-world OOD distributions in the dataset. We experimented with different K values (K= $\{2, 5, 10, 20\}$) and found that K=5 (reported in Table~\ref{tab:res_table}), close to the average number of unknown class instances per frame, performed best across all metrics. Overall, the remaining gap between the two-stage oracle MLP (GT Unknowns in Table~\ref{tab:res_table}) and this highest-performing unknown generation method suggests that future works should focus on improving this pseudo-labelling process.



\section{CONCLUSIONS}

In this work, we presented a new two-stage method for detecting unknown objects called HD-OOD3D. By exploiting a sufficiently large dataset such as Argoverse 2 to train the unknowns in a supervised manner, we demonstrated the superiority of two-stage methods over single-stage methods, particularly our proposed HD-OOD3D, which utilizes more input representations for better object categorization. We then broke down the classic evaluation protocol for unknown objects to highlight the importance of certain hyperparameter choices for a fair assessment of methods. Finally, scaling the learning of unknown objects can only be achieved through unsupervised training. We compared different approaches for generating unknowns by using only the annotations of known objects from Argoverse 2. We showed that top-K auto-labelling offers more promising performance perspectives compared to resizing. Ultimately, the task of detecting unknown objects remains a challenging one, with current performance far from what would enable deployment on robotics platforms, making it a field that still requires further research to reach maturity.


\bibliographystyle{IEEEtran}
\bibliography{iros_main}




\end{document}